\documentclass[10pt,conference]{IEEEtran}
\usepackage{textcomp}
\usepackage[table]{xcolor}
\usepackage{booktabs}
\usepackage{xspace}
\usepackage[utf8]{inputenc}
\DeclareUnicodeCharacter{FB01}{fi}
\usepackage{subcaption}
\usepackage{amssymb}
\usepackage{pifont}

\def\BibTeX{{\rm B\kern-.05em{\sc i\kern-.025em b}\kern-.08em
    T\kern-.1667em\lower.7ex\hbox{E}\kern-.125emX}}

\usepackage{listings}
\usepackage{subcaption}
\usepackage{dblfloatfix}
\usepackage{amsmath}
\usepackage{amsthm}
\usepackage{amssymb}
\usepackage{accents}
\usepackage{algorithm}
\usepackage{algorithmicx}
\usepackage{algpseudocode}
\usepackage{url}
\usepackage{color}
\usepackage{wrapfig}

\usepackage{pgf}
\usepackage{tikz}
\usetikzlibrary{calc}
\usetikzlibrary{shapes}
\usetikzlibrary{arrows}
\usepackage{pgfplots}
\usepackage{pgfplotstable}
\pgfplotsset{compat=1.13}

\newcommand{\ignore}[1]{}

\newcommand\x[1]{\ensuremath{\mathit{#1}}}

\definecolor{mygreen}{rgb}{0,0.6,0}
\definecolor{mygray}{rgb}{0.5,0.5,0.5}
\definecolor{mymauve}{rgb}{0.58,0,0.82}

\def\x{{\mathbf x}}

\newcommand\planet{{\sl Planet}\xspace}
\newcommand\reluplex{{\sl ReLuplex}\xspace}
\newcommand\eran{{\sl ERAN}\xspace}
\newcommand\neurify{{\sl Neurify}\xspace}

\newcommand\dave{{DAVE-2}\xspace}
\newcommand\dronet{{DroNet}\xspace}

\lstdefinelanguage{toml}{
basicstyle=\scriptsize\ttfamily,
sensitive=false,
morecomment=[f][\bf][0]{[},
showstringspaces=false,
}

\usepackage{hyperref}
\hypersetup{hidelinks}

\newcommand{\approach}{R4V\xspace}

\newcommand{\xmark}{\ding{55}}
\newcommand{\cmark}{\ding{51}}

\begin{document}

\title{Refactoring Neural Networks for Verification}

\author{\IEEEauthorblockN{David Shriver, Dong Xu, Sebastian Elbaum, Matthew B. Dwyer}
	\IEEEauthorblockA{\textit{Department of Computer Science, University of Virginia} \\
		\texttt{\{dlshriver, dx3yy, selbaum, matthewbdwyer\}@virginia.edu}}
}

\maketitle

\begin{abstract}
	Deep neural networks (DNN) are growing in capability and applicability.
	Their effectiveness
	has led to their use in safety critical and autonomous systems,
	yet there is a dearth of cost-effective methods available for reasoning
	about the behavior of a DNN.
		
	In this paper, we seek to expand the applicability and scalability
	of existing DNN verification techniques through DNN
	refactoring.  A DNN refactoring defines (a) the transformation of
	the DNN's architecture, i.e., the number and size of its layers,
	and (b) the distillation of the learned relationships
	between the input features and function outputs of the original to
	train the transformed network.
	Unlike with traditional code refactoring, DNN refactoring does
	not guarantee functional equivalence of the two networks, but
	rather it aims to preserve the accuracy of the original network
	while producing a  simpler network that is amenable to more
	efficient property verification.
		
	We present an automated framework for DNN refactoring,
	and demonstrate its potential effectiveness through three
	case studies on networks used in autonomous systems.
\end{abstract}


\section{Introduction}\label{sec:introduction}

Deep learning has emerged over the past decade as a
means of synthesizing implementations of complex
classification \cite{krizhevsky-etal:nips:2012},
clustering~\cite{duda-etal:2001:book}, and control functions~\cite{mnih-etal:corr:2013,mnih-etal:Nature:2015}.
These implementations, in the form of Deep Neural Networks (DNN),
offer the potential for new or enhanced system capability.
For example, functions that compute control outputs from
bit-mapped images have been synthesized that can outperform
a trained human on a range of complex ``game playing'' tasks~\cite{mnih-etal:corr:2013,mnih-etal:Nature:2015}
and those techniques have been adapted
to replace heuristic and human control in autonomous
systems~\cite{7838739,5611206}.

As these techniques mature they are being
deployed in safety critical systems. However,
there is a lack of cost-effective techniques for reasoning
about the behavior of DNNs which, in turn, limits the development of
safety arguments for such systems and is a cause for concern, e.g.,
automotive industry standards are under scrutiny for
not addressing DNNs~\cite{2018-01-1075}.
To address this need, researchers have been actively exploring
algorithms for verifying that the behavior of a trained DNN
meets explicitly stated correctness properties.
\reluplex~\cite{katz-etal:CAV:2017} and \planet~\cite{ehlers:ATVA:2017}
were the first such tools, but 
more than 18 DNN verification algorithms~\cite{
8318388,8418593,DBLP:conf/ijcai/RuanHK18,NIPS2018_8278,tjeng2018evaluating,Bastani:2016:MNN:3157382.3157391,Dvijotham18,DBLP:conf/icml/WongK18,DBLP:conf/iclr/RaghunathanSL18,DBLP:conf/nips/WangPWYJ18,DBLP:conf/uss/WangPWYJ18,DBLP:conf/icml/WengZCSHDBD18,DBLP:conf/cav/HuangKWW17,Boopathy2019cnncert,DBLP:conf/nfm/DuttaJST18,DBLP:conf/nips/BunelTTKM18}
have been developed in the past two years.

Despite this progress, the state-of-the-art in DNN verification
is far from adequate for addressing the scale and complexity of DNNs
used in autonomous systems.
As we demonstrate in \S\ref{sec:study},
networks such as DAVE-2~\cite{bojarski-etal:corr:2016:DAVE2} and 
DroNet~\cite{loquercio-etal:RAL:2018:dronet},
which are used in autonomous driving and autonomous UAV navigation,
are too complex for cost-effective verification.

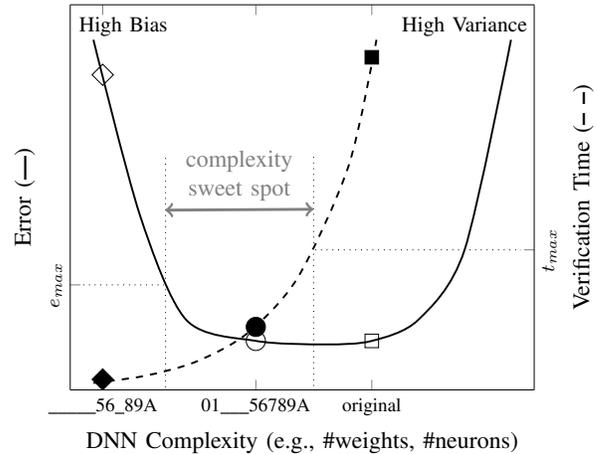
\begin{figure}
\centering
\pgfdeclarelayer{background}
\pgfsetlayers{background,main}
\begin{tikzpicture}[scale=0.9]

  \begin{axis}[
    axis y line*=left,
    ylabel={Error (\textbf{---})},
    ytick={0.3},
    yticklabel style={rotate=90},
    yticklabels={\footnotesize $e_{max}$},
    ymin=0.0,
    xlabel={DNN Complexity (e.g., \#weights, \#neurons)},
    xtick={1.2,4.5,7},
    xticklabels={
      {\footnotesize \_\_\_\_\_56\_89A},
      {\footnotesize 01\_\_\_56789A},
      {\footnotesize original}},
    xmin=0.5,
    xmax=10.5
    ]
    \addplot+[thick,smooth,color=black,mark=none] coordinates
      {(1,1) (2,0.5) (3,0.2) (4.5,0.14) (6,0.13) (7,0.14) (8,0.2) (9,0.4) (10,1) };

    \node[align=center] at ($(axis cs:1.5,1)+(1mm,2mm)$) {\small High Bias};
    \node[align=center] at ($(axis cs:9,1)+(0mm,2mm)$) {\small High Variance};

    \node (emaxl) at (axis cs:0.6,0.3) {};
    \node (emaxr) at (axis cs:2.7,0.3) {};
    \draw[-, dotted] (emaxl) -- (emaxr);

    \node (emaxt) at ($(emaxr)+(-1mm,20mm)$) {};
    \node (emaxb) at ($(emaxr)+(-1mm,-20mm)$) {};
    \draw[-, dotted] (emaxt) -- (emaxb);

    \node[draw, circle, inner sep=1mm] (justright) at (axis cs:4.5,0.14) {}; 
    \node[draw, circle, fill=black, inner sep=1mm] at (axis cs:4.5,0.18) {}; 

    \node[draw, diamond, inner sep=0.8mm] (toomucherror) at (axis cs:1.2,0.9) {}; 
    \node[draw, diamond, fill=black, inner sep=0.8mm] at (axis cs:1.2,0.03) {}; 

    \node[draw, rectangle, inner sep=1mm] at (axis cs:7,0.14) {}; 
    \node[draw, rectangle, fill=black, inner sep=1mm] (toocostly) at (axis cs:7,0.95) {}; 

  \end{axis}

  \begin{axis}[
    axis y line*=right,
    axis x line=none,
    ylabel={Verification Time (\textbf{-- --})},
    ytick={0.4},
    yticklabel style={rotate=90},
    yticklabels={\footnotesize $t_{max}$},
    ymin=0.0,
    xmin=0.5,
    xmax=10.5]
    \addplot+[thick,smooth,dashed,color=black,mark=none] coordinates
      {(1,0.02) (2.1,0.04) (3.2,0.08) (4.3,0.16) (5.4,0.32) (6.5,0.64) (7.1,1) };
    \node (tmaxl) at (axis cs:5.6,0.4) {};
    \node (tmaxr) at (axis cs:10.4,0.4) {};
    \draw[-, dotted] (tmaxl) -- (tmaxr);

    \node (tmaxt) at ($(tmaxl)+(1mm,15mm)$) {};
    \node (tmaxb) at ($(tmaxl)+(1mm,-25mm)$) {};
    \draw[-, dotted] (tmaxt) -- (tmaxb);

    \draw[very thick,gray,<->] ($(emaxt)+(-1mm,-10mm)$) -- node[midway,above,align=center] {complexity\\sweet spot} ($(tmaxt)+(0mm,-9mm)$);

  \end{axis}

\end{tikzpicture}
\caption{Conceptual diagram depicting (a) the bias-variance tradeoff in deep-learning and (b) the growth in verification time with model complexity.}
\label{fig:error-verifiability}
\end{figure}

{\bf Balancing Error and Verifiability}:
Fig.~\ref{fig:error-verifiability} illustrates, with the solid curve,
a conceptual plot of the classic bias-variance tradeoff in training
a machine learning model~\cite{geman-etal/NC/1992}.
While there are many possible measures of complexity of a machine learning model,
for training a DNN one can understand complexity in terms of the number of
parameters, or weights, that are learned.
This plot captures the intuition that:
(a) low complexity networks have insufficient capacity
to fit the training data (high-bias), and 
(b) high complexity networks risk overfitting the training data (high-variance).
The plot also reflects that, in practice, DNNs tend not to overfit until their complexity becomes
very large relative to the training data, e.g., \cite{zhou-etal/ICLR/2019,neyshabur-etal/ICLR/2019,cheng-etal/SPM/2018}.
This has the effect of extending the region of low error far 
to the right in terms of Fig.~\ref{fig:error-verifiability}.

For developers who are focused exclusively on low 
error (high accuracy) this means that there is little disincentive
to adding neurons, and thereby additional parameters, to the DNN
in the hopes of achieving lower error.
However, for developers who seek both low error and the ability to verify
properties, increasing DNN complexity, e.g., by adding neurons,  
can substantially increase verification time.
Fig.~\ref{fig:error-verifiability} illustrates, with the dashed curve,
a conceptual plot of the growth in verification time
with increasing DNN complexity -- 
Fig.~\ref{fig:cs3} in \S\ref{sec:study} reports data
substantiating the growth in verification time with the number of neurons.

\begin{figure*}[th]
\centering
\includegraphics[width=\textwidth]{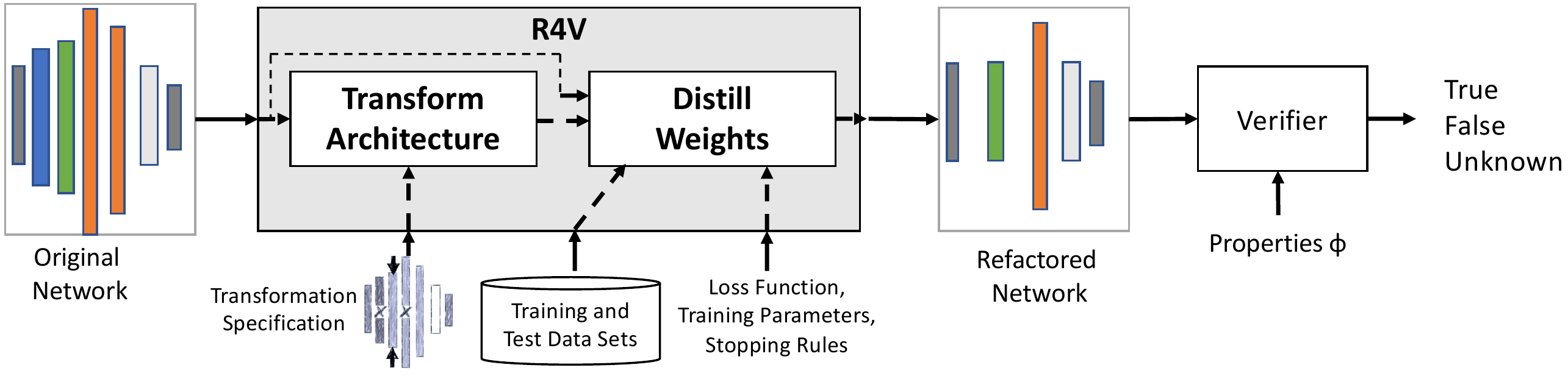}
\caption{The \approach Approach}
\label{fig:basicapproach}
\end{figure*}

Consider a developer who needs a DNN with error
less than $e_{max}$ and who is willing to wait $t_{max}$ for
verification results.
Conceptually, these constraints define a lower bound on DNN
complexity, i.e., the complexity necessary to achieve lower error than $e_{max}$,
and an upper bound on DNN complexity, i.e., the complexity beyond
which verification time exceeds $t_{max}$.
These bounds define a \textit{complexity sweet spot} within which
a DNN is both accurate and verifiable.
In this paper, we propose automated support that developers
can use to navigate error-verifiability tradeoffs resulting from DNN 
complexity to produce a network within this sweet spot.

{\bf Refactoring for Verification}: To achieve this 
we adapt the concept of 
\textit{refactoring} to deep neural networks.
Code refactoring~\cite{fowler2018refactoring} typically seeks to
(a) restructure a software system to facilitate subsequent development
activities, while
(b) preserving the behavior of the original software system. 
Fowler's original focus was on improving maintainability and
extensibility, but researchers have adapted the concept in myriad ways.
For example, 
to enhance the testability~\cite{cinneide2011automated,harman2011refactoring}
and verification~\cite{yin2009exploiting} of software.

DNN \textit{refactoring for verification} (\approach)
applies and adapts the basic principles of code refactoring.
As depicted in Fig.~\ref{fig:basicapproach},
given an original trained DNN, refactoring proceeds in two phases: 
architectural transformation and weight distillation.

The architecture of a new DNN is defined as a transformation
of the original network's architecture based on 
a developer provided \textit{transformation specification}. 
As described in \S\ref{sec:approach}, 
a specification designates layers to drop, layers to be downscaled
to eliminate neurons, and layers to be transformed to meet verifier 
requirements.  DNN transformation proceeds by propagating 
inter-layer constraints
to produce a well-defined, but untrained, DNN that is less complex
than the original and thereby facilitates verification.

This transformed DNN architecture is trained using 
knowledge distillation which seeks to train the transformed
DNN to match the accuracy of the original, trained 
DNN~\cite{hinton-etal:distillation}.
Like traditional training, distillation requires the specification
of training and test data sets and training parameters, such as 
the specific loss function and stopping rules.

Unlike code refactoring, where exact preservation of observable behavior
is often a goal, the stochastic nature of DNN training makes this difficult.
However, for many applications the behavior of a DNN is captured by
its test error.
Since the distillation process drives training of the transformed DNN
to minimize its difference with the original, it preserves
behavior by trying to match the error of the networks.

{\bf \approach in Action}:
Fig.~\ref{fig:error-verifiability} illustrates how DNN refactoring can
help the developers of the DAVE-2 network verify properties and
achieve desired error using a binary-search on the complexity
of refactored DNNs.  We illustrate this by referencing data on
refactored versions of DAVE-2 reported in Fig.~\ref{fig:cs3}.

The original DAVE-2 network computes steering angles from images using
a 13 layer DNN comprised of 82669 neurons.  
Unfortunately, on average it takes 20,000 seconds 
to run the state-of-the-art \neurify\cite{DBLP:conf/nips/WangPWYJ18} verifier to check
a single property and even then the verifier returns
an \textit{unknown} result -- it is unable to prove or disprove
property conformance.
Fig~\ref{fig:error-verifiability} depicts the error of the
network and its verification time with an unfilled square and a 
filled square, respectively.
The high-cost of verifying this network means it is too complex and we
should refactor it to reduce its complexity.

An extreme aggresive refactoring might drop 6 of the layers
resulting in a very small DNN with only 161 neurons, denoted
{\_\_\_\_\_56\_89A}\footnote{This DNN naming scheme 
uses ``\_'' to denote
a dropped layer -- the original network would be denoted
{0123456789A}.  Since DNN refactorings 
always preserve the input and output layers we do not include them 
in the name.}.
This refactored DNN is easy to verify, taking only 30
seconds on average, which we depict as a filled diamond.
Unfortunately, this network lacks sufficient complexity to match the training
data.  Its error, which we denote with a diamond, has an MSE of 0.062 relative to the original network -- this network has not preserved the original
networks behavior.
Clearly refactoring has swung too far towards low DNN complexity.

Directing refactoring back towards greater complexity splits the difference
between the first two networks by 
dropping only fewer layers. 
The resulting 8 layer DNN, {01\_\_\_56789A}, has 74045 neurons.
On average, properties of this network can be verified in
5000 seconds and it achieves a relative Mean Squared Error (MSE) of 0.008.
If $t_{max}$ and $e_{max}$ are 3 hours and a relative MSE of 0.01 then, 
as depicted by the filled and unfilled circles, this refactored network would 
lie within the complexity sweet spot.
The ability to cost-effectively verify
DNN {01\_\_\_56789A}, allows developers to determine
whether it meets critical properties and, if so, to deploy since
it has acceptably low error.

The contributions of this work lie in 
(1) identifying the notion of a complexity sweet spot that
allows developers to balance DNN error and verifiability;
(2) developing automated DNN refactoring techniques that can
be used to side-step the limitations of existing DNN verifiers
and adjust DNN complexity to meet desired error and verification constraints;
and
(3) demonstrating the potential of this approach on a series
of case studies using two substantial
DNNs deployed in autonomous systems and four state-of-the-art
DNN verifiers.

\section{Background}\label{sec:overview}
In this section, we give background on DNNs, DNN verification, and
DNN distillation as a prelude to describing \approach.

\subsection{Deep Neural Networks}
A deep neural network $\mathcal{N}$ encodes an approximation of a target function $f: \mathbb{R}^{n} \rightarrow \mathbb{R}^{m}$.
DNNs are comprised of a sequence of layers, $l_0, l_1, \ldots, l_k$,
where $l_0$ has an input domain of $\mathbb{R}^{n}$, $l_k$
has an output domain of $\mathbb{R}^m$, and
the output domain of $l_{i}$ is the same as the input domain of $l_{i+1}$.
Each layer defines an independent function and the overall network's
function approximation is their composition $l_k \circ l_{k-1} \ldots \circ l_0$.
As discussed below, layers may vary in their internal structure, but all are comprised
of a set of \textit{neurons} that accumulate a weighted sum of other neurons
and then
apply an \textit{activation function} (e.g., ReLU, Sigmoid) to the neuron output to
introduce non-linearity.

The literature on machine learning has developed a broad range of
\textit{layer type}s and explored the benefits of different layer combinations
in realizing accurate approximations of different target functions, e.g., \cite{Goodfellow-et-al-2016}.
For instance, layer types have been defined that specialize in processing
input image data, e.g.,
convolutional and max pooling,
in providing flexible expressive power for function approximation,
e.g., fully-connected, and
in producing output classifications, e.g., softmax.
A layer is defined by instantiating a layer type with parameter
definitions.
For instance, a convolutional
layer type might define the
size of its input, the number of output channels, its kernel size, and its stride.

Layers can also have a more complex structure, with branches of computation
that are recombined to produce the layer output. For example, residual blocks,
e.g., \cite{he-etal:CVPR:2016:resnet}, have 2 computation paths. One path applies
a set of computations, such as convolution and batch normalization, while the other
path is an identity function, mapping the input to the final recombination operation.
The two paths are combined by adding their outputs.
In this work we represent these complex blocks as $block(C, S)$, where $C$ is the
recombination operation (e.g. $Add$) and $S$ is the set of computation paths.
Residual blocks can be represented using this notation as
$block(Add, \lbrace I, f \rbrace )$, where $I$ is the identity operation, and $f$
is the computation to apply to the computation path.

The definition of a set of layers and their interconnections
defines the \textit{architecture} of the network.
An architecture is \textit{trained} using a population of
labeled training data, $D$, consisting of training instances, $(\x,y)$.
Training instances are evaluated, $\mathcal{N}(\x)$,
and the \textit{loss} relative to the expected output, $y$, is used
to drive a gradient-descent process, called
\textit{backpropagation}, which updates the network \textit{weight}s based
on their degree of contribution to the loss.
The term \textit{training parameters} refers collectively to
the set of choices made in instantiating this training process, including:
how the $(\x,y)$ are sampled from $D$, the loss function
(e.g., mean squared error, cross entropy), and when to stop training.

The \textit{accuracy} of the network estimates the degree to which
$\mathcal{N}(\x) \approx f(\x)$ by computing the percentage
of labels that are matched by $\mathcal{N}$ on a set of labeled instances
that are disjoint from $D$.
It can be measured to determine the progress of training,
e.g., validation accuracy,
or to determine generalization of the network after training is
completed, e.g., test accuracy.
The \textit{error} of a network is 1 minus its accuracy.

\subsection{DNN Verification}
Given a network $\mathcal{N}$ and a correctness property, $\phi$, that defines
a set of constraints over the inputs $\phi_{x}$, and a set of constraints over the outputs $\phi_{y}$, verification of $\mathcal{N}$ seeks to prove:
$\forall{x\in\mathbb{R}^{n}}\;\phi_{x}(x) \Rightarrow \phi_{y}(\mathcal{N}(x))$.

A widely studied class of properties express the \textit{local robustness}
of a DNN.
Such properties state that inputs within an $\epsilon$-ball centered at
$\x$ should be mapped to the same value,
$\forall{\delta \in [0,\epsilon]^n}\ \mathcal{N}(\x) = \mathcal{N}(\x\pm\delta)$, where $n$ is the dimension of $\x$.
For example, a network that classifies road signs should still classify an image
of a stop sign correctly, even after a small amount of noise is applied to the
image.

\begin{figure}
\centering
\pgfdeclarelayer{background}
\pgfdeclarelayer{wrap}
\pgfsetlayers{background,wrap,main}
\begin{tikzpicture}[
phase/.style={rectangle, draw, dashed, inner sep=0pt, fit=#1}
]
  \node[name=teacher, fill=white, draw, align=center] {\textit{teacher}\\ ~\\ Original\\ DNN \\~};

  \node[name=student, fill=white, draw, right of=teacher, xshift=30mm, align=center] {\textit{student}\\ ~\\ Refactored\\ DNN \\~};

  \node[name=loss, fill=white, draw, above of=teacher, yshift=5mm, xshift=20mm, align=center] {~~loss~~};

  \node[name=labelled, below of=loss, yshift=-20mm, xshift=0mm, align=center] {$(\x,y) \in D$};

  \draw[->,dashed, thick] ($(labelled.north)+(-2mm,0mm)$) -- node[left] {$y$} ($(loss.south)+(-2mm,0mm)$);

  \draw[->,dashed, thick] (teacher.north) |- node[above] {$\overrightarrow{o_t(\x)}$} (loss.west);
  \draw[->,dashed, thick] (student.north) |- node[above] {$s(\x), \overrightarrow{o_s(\x)}$} (loss.east);

  \draw[->,dotted, thick] ($(loss.south)+(2mm,0mm)$) |- node[right, xshift=-1mm, yshift=-5mm, align=center] {back\\prop.} (student.west);

  \draw[->,thick] (labelled.east) node[above,xshift=4mm] {$\x$} -| (student.south);
  \draw[->,thick] (labelled.west) node[above,xshift=-4mm] {$\x$} -| (teacher.south);

\end{tikzpicture}
\caption{DNN Distillation}
\label{fig:distillation}
\end{figure}
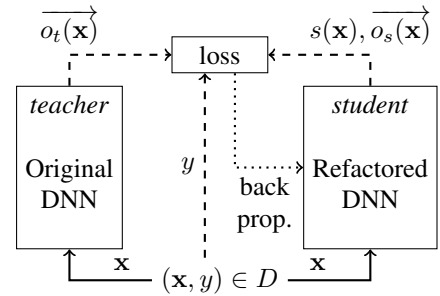

A recent survey~\cite{DBLP:journals/corr/abs-1903-06758}
describes a variety of algorithmic approaches for verifiying
such properties.
Reachability methods that calculate approximations of the sets
of values that can be
computed as output of the network given an input constraint
include: MaxSens~\cite{8318388},
Ai2~\cite{8418593}, DeepGo~\cite{DBLP:conf/ijcai/RuanHK18}, and
\eran~\cite{NIPS2018_8278,singh-etal:POPL:2019:deeppoly,singh-etal/ICLR/2019}.
Methods that formulate property violations as a threshold for
an objective function and
use optimization algorithms to determine failure to meet that threshold include:
MIPVerify~\cite{tjeng2018evaluating},
ILP~\cite{Bastani:2016:MNN:3157382.3157391},
Duality~\cite{Dvijotham18},
ConvDual~\cite{DBLP:conf/icml/WongK18}, and
Certify~\cite{DBLP:conf/iclr/RaghunathanSL18}.
Search methods that explore regions of the input space where they then
formulate reachability or optimization sub-problems include:
\neurify~\cite{DBLP:conf/nips/WangPWYJ18},
ReluVal~\cite{DBLP:conf/uss/WangPWYJ18},
Fast-lin\& Fast-lip~\cite{DBLP:conf/icml/WengZCSHDBD18},
DLV~\cite{DBLP:conf/cav/HuangKWW17}, CROWN~\cite{zhang2018crown}, CNN-CERT~\cite{Boopathy2019cnncert},
Sherlock~\cite{DBLP:conf/nfm/DuttaJST18},
Bab~\cite{DBLP:conf/nips/BunelTTKM18},
\reluplex~\cite{katz-etal:CAV:2017}, and \planet~\cite{ehlers:ATVA:2017}.
In \S\ref{sec:study} we report on the use of four of these DNN verifiers in
combination with \approach.

\begin{figure*}[th!]
    \centering
    \begin{minipage}[t]{0.18\linewidth}
        \includegraphics[width=\linewidth]{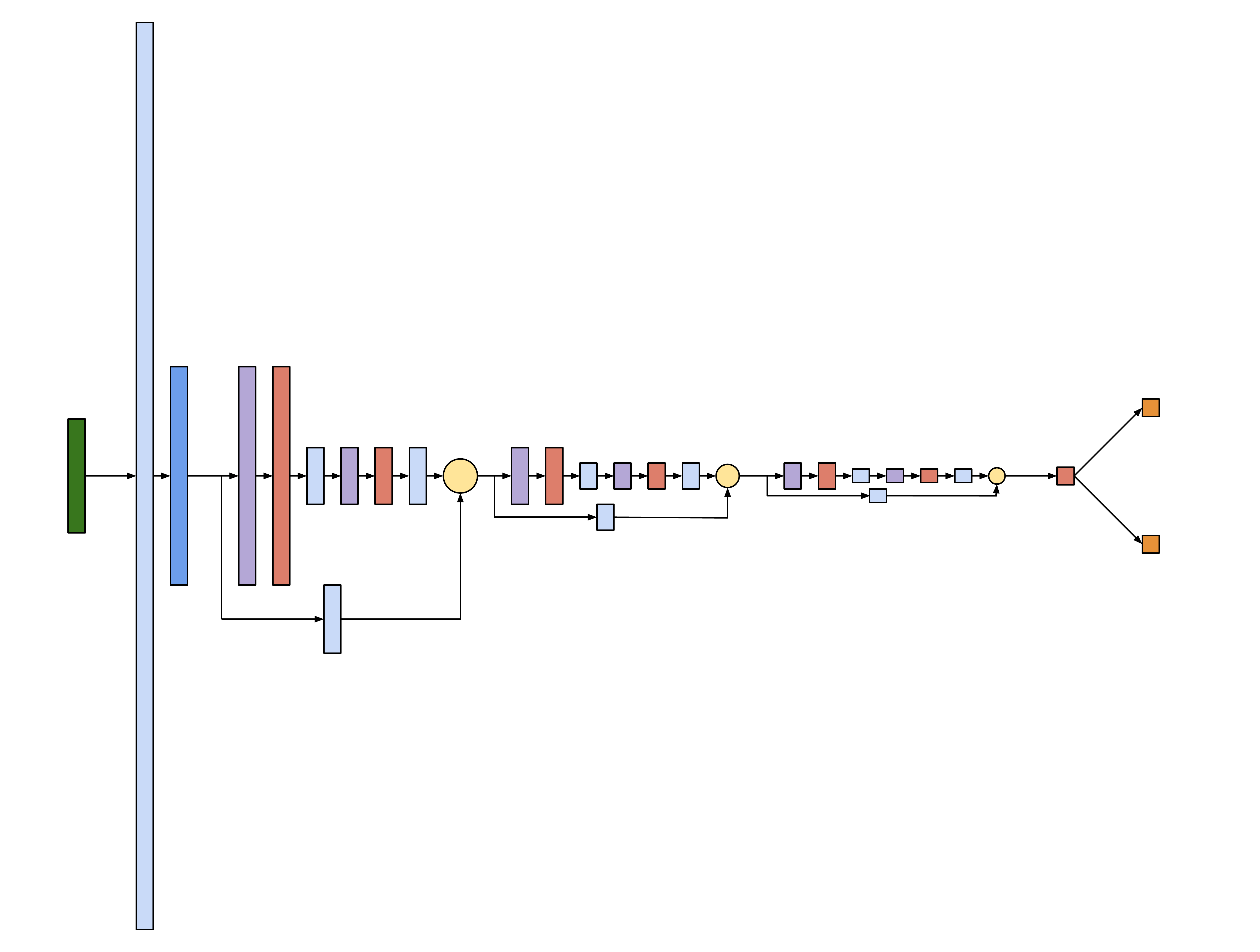}
        \center{Original \dronet}
    \end{minipage}%
    \begin{minipage}[t]{0.18\linewidth}
        \includegraphics[width=\linewidth]{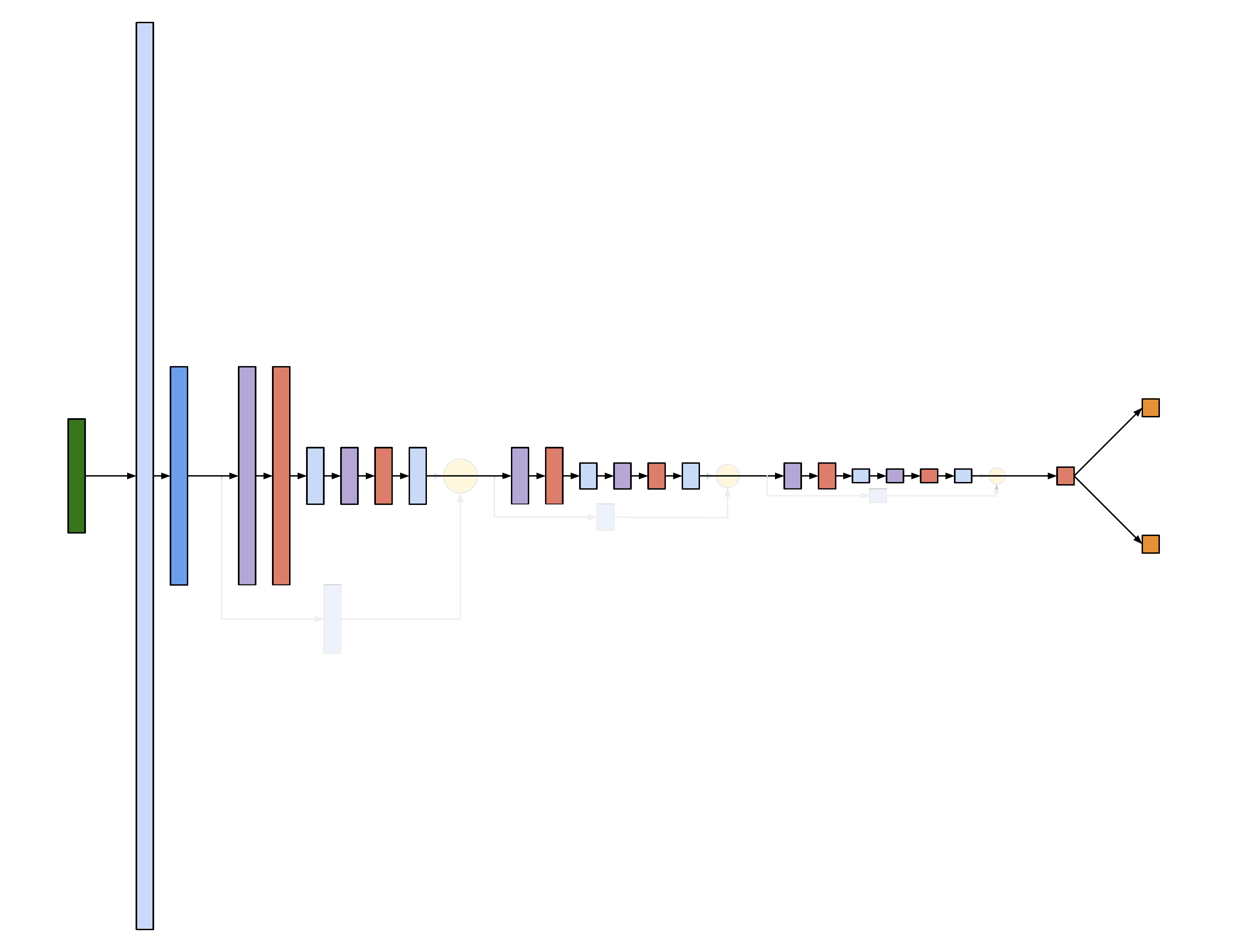}
        \center{\textit{linearize}}
    \end{minipage}%
    \begin{minipage}[t]{0.18\linewidth}
        \includegraphics[width=\linewidth]{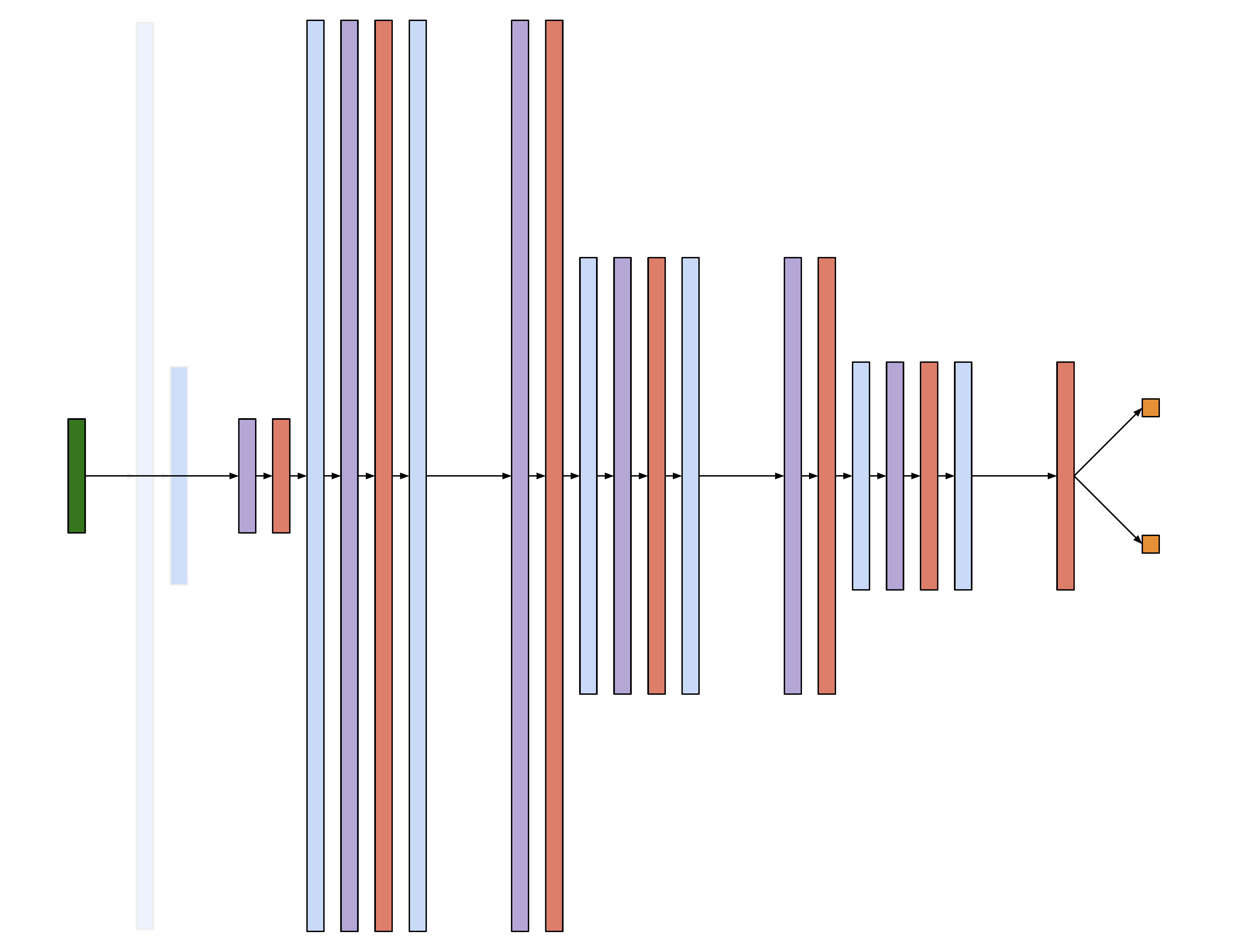}
        \center{\textit{drop(0)}}
    \end{minipage}%
    \begin{minipage}[t]{0.18\linewidth}
        \includegraphics[width=\linewidth]{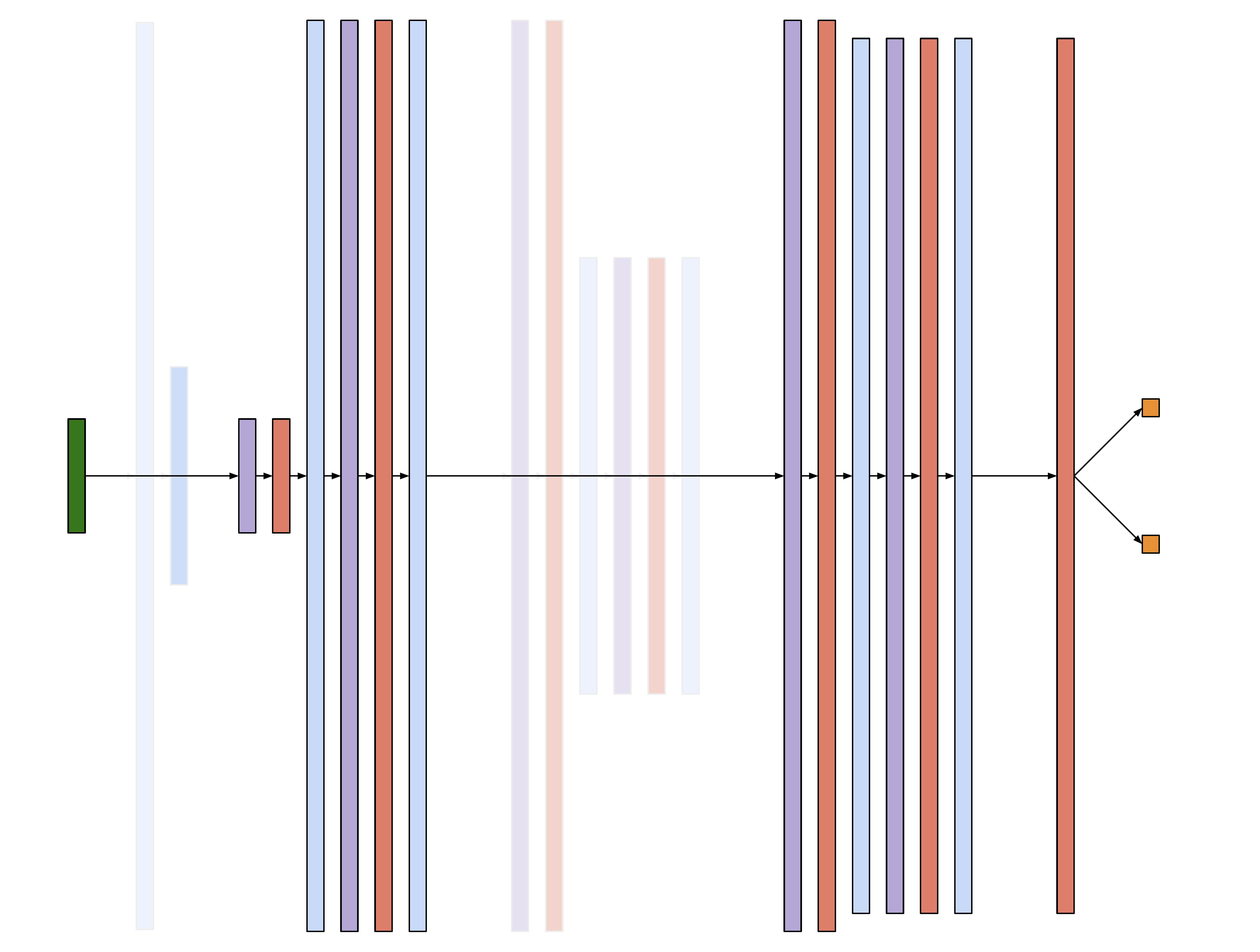}
        \center{\textit{drop(2)}}
    \end{minipage}%
    \begin{minipage}[t]{0.18\linewidth}
        \includegraphics[width=\linewidth]{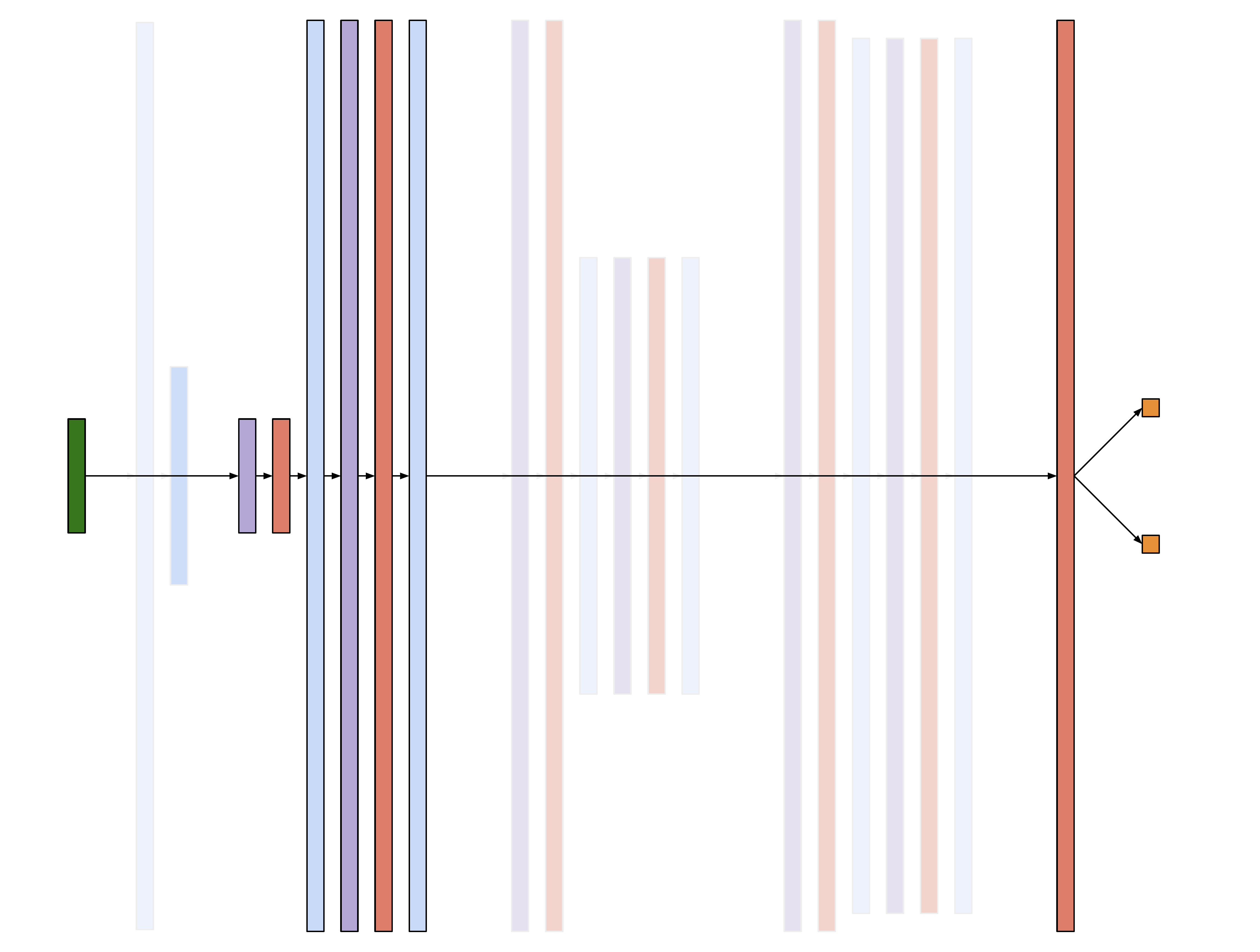}
        \center{\textit{drop(3)}}
    \end{minipage}%
    \begin{minipage}[t]{0.085\linewidth}
        \includegraphics[width=\linewidth]{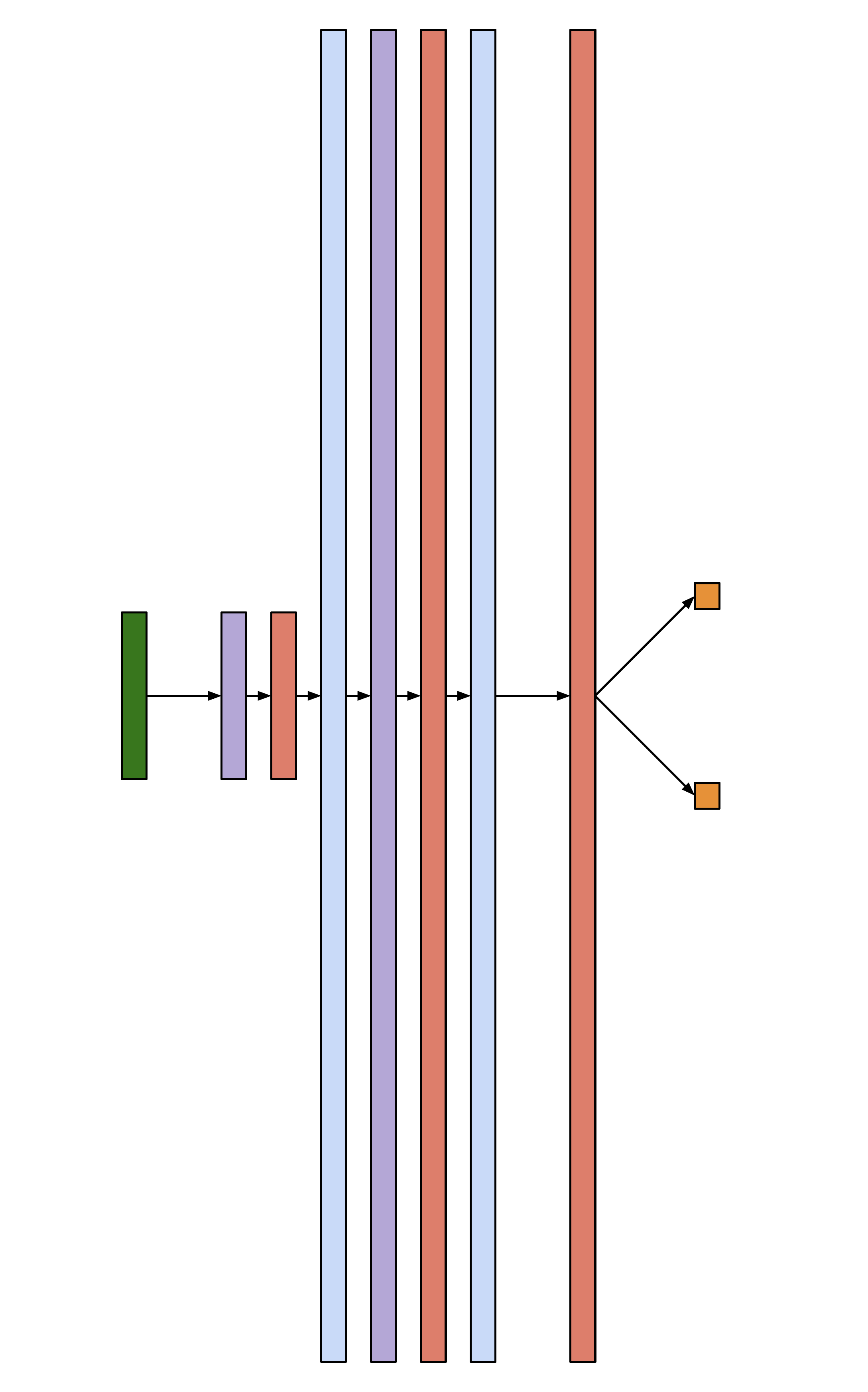}
        \center{Refactored \dronet}
    \end{minipage}
    \caption{Refactoring the \dronet DNN.}
    \label{fig:dronetrefactoring}
\end{figure*}

\subsection{Network Distillation}
Hinton et al. \cite{hinton-etal:distillation} introduced
\textit{knowledge distillation} as a general approach to addressing the differing
requirements in training and deploying DNNs (e.g., less computation, lower energy consumption).
Distillation has
proven useful in addressing varied deployment requirements~\cite{romero-etal/ICLR/2015,chen-etal/NIPS/2017,howard-etal/corr/2017}.
In \approach, we employ distillation
to address the requirement that DNNs be verified
prior to deployment in safety critical systems.

As depicted in Fig.~\ref{fig:distillation}, distillation trains
a \textit{student} DNN to match the accuracy of an trained
\textit{teacher} DNN.
Training inputs, $\x$, are fed to each network.
In traditional training, loss would be computed between
the label $y$ -- a \textit{hard} training target -- and the
output inferred by the network being trained, $s(\x)$.
Distillation can also include differences in
\textit{soft} training targets in computing loss.
Soft targets are vectors of values computed in the output layers of the
networks, $\overrightarrow{o_t(\x)}$ and $\overrightarrow{o_s(\x)}$;
for classification networks these would be the normalized logits in
the softmax layer.
The computed loss drives backpropagation to update the
student's weights, but the addition of soft targets can better help
the student match the accuracy of the teacher.

\section{Approach}\label{sec:approach}
The goal of our approach, \approach, is to increase
the applicability and scalability of verification
through refactoring -- transferring the knowledge
of a complex DNN into a smaller simpler network that is more
amenable to verification.
In this section we discuss the components of \approach, as
outlined in Fig.~\ref{fig:basicapproach}, describe its implementation,
and  present recommendations on how to systematize
its application.

\subsection{Transforming DNN Architecture }
Different strategies can be applied to reduce the complexity of a DNN.
We define a simple set of transformations over DNN layers. These transformations permit
dropping layers, scaling layers, and changing certain types of layers that are not
supported by verification tools. We also define a utility operation to apply transformations
to    layers  that satisfy a given predicate.  

\textbf{Drop layers}:
The drop layer operation (Eq.~\ref{eq:droplayer}) takes in a set of layer indices $L$, and removes those layers from the DNN.
\begin{align}\label{eq:droplayer}
    drop(L):\, & \forall{l \in L}\;{l' = \bot}
\end{align}
After dropping a layer, the output shape of the preceding layer is used to update
successive layers, to ensure that the new architecture is valid.
The drop operation can be applied to layers that perform some computation
on their input. We prohibit its application to the output layer to
ensure that the transformed architecture produces outputs with the same
dimensions as the original architecture. We also prohibit dropping reshaping
layers, such as \textit{Transpose} and \textit{Flatten}, since they  are
required to ensure that the input to a layer are the correct shape.

\textbf{Scale layers}:
The scale layer operation (Eq.~\ref{eq:scalelayer}) takes in a set of layer indices $L$, and a scale factor $f$,
and scales the number of neurons in each specified layer by the factor $f$.
\begin{align}\label{eq:scalelayer}
    scale(L, f):\, & \forall{l \in L}\;{(l \not= \bot) \rightarrow (\#l' = \lfloor{f \cdot \#l}\rfloor)}
\end{align}
where $\#$ denotes the size of a layer.
After scaling a layer, its new output shape is used to update the parameterization
of successive layers, to ensure that the new architecture is valid.
Our current framework supports  the scale operation on the input layer, fully-connected
layers, and convolutional layers. Fully-connected layers
are scaled by changing the number of neurons in the layer. Convolutional
layers are scaled by changing the number of kernels. The input layer
is scaled by its height and width dimensions.

\textbf{Linearize layers}:
In general, residual blocks have one path that performs a computation on the input, and
a second path that performs the identity operation, or resizes the input to match the
output shape of the computation path.
The linearize operation removes the non-computation path from the residual block so that
the layer involves a single path.
It (Eq.~\ref{eq:linearizelayer}) takes a set of indices $L$ of residual layers, and removes
the identity connection from each.
\begin{align}\label{eq:linearizelayer}
    linearize(L):\, & \forall{l \in L}\;{l = block(Add, \lbrace I, f \rbrace) \rightarrow l' = f}
\end{align}

\textbf{Forall layers}:
We also define a utility operation $\mathit{forall}$ (Eq.~\ref{eq:foralllayers}), that applies
a partially parameterized operation $\lambda$ over all layers of a DNN $\mathcal{N}$,
that satisfy predicate $\varphi$.
\begin{align}\label{eq:foralllayers}
    \mathit{forall}(\mathcal{N}, \varphi, \lambda) & = \lambda(\lbrace l \in \mathcal{N} \;|\; \varphi(l) \rbrace)
\end{align}

Fig.~\ref{fig:dronetrefactoring} illustrates the individual transformations
resulting from this composition of operators:
\begin{align*}
\mathit{forall}( & \mathit{forall}(\mathrm{\dronet}, \mathit{isResidual}, \mathit{linearize}), \\
& \mathit{isLayer}(\{0,2,3\}), \mathit{drop})
\end{align*}
where \textit{isResidual} and \textit{isLayer} are predicates that
determine whether a layer is residual
and whether a layer's identifier is in a specified set, respectively.
The colors in the figure depict different layer types and their height
corresponds to the number of neurons.
The transformation is non-trivial, because whenever a layer's shape
is modified the adjacent layers have to be updated.  

We have applied and investigated these operations 
individually and in combination over several networks, including
the ones that we report on in \S\ref{sec:study}.  
We have considered extending the transformation framework 
with additional capabilities,
for example, to \textit{add} a layer.
To date, however, we have focused on capabilities that reduce the complexity
of refactored networks -- adding layers works against that goal -- and
that are necessary to support existing DNN verifiers.

\subsection{Distilling Weights in the Transformed DNN}
We employ knowledge distillation to refactor DNNs
that retain accuracy relative to the original network.
Since distillation is a flexible, highly parameterized,
DNN cross-training framework, our approach identifies three
degrees of freedom in adapting distillation for verification.

\noindent{\bf Datasets.} These nominally include training, validation, and test sets.
Usually, the sets used on the teacher will be used on the student as well. However,
\approach supports the utilization of different sets that may be needed to, for example,
accelerate distillation or performing additional checks on external representativeness. 

\noindent{\bf Training parameters.}
As with other DNN training approaches, there is a large hyperparameter space
for tuning performance.  Our approach has been to adopt the training parameters
of the original network as a starting point.  
For refactoring classification networks, 
we follow prior work beginning with Hinton et al.~\cite{hinton-etal:distillation},
and incorporate soft targets, but hard targets can be used to provide 
additional information for computing loss.
For refactoring regression networks, we distill strictly using the 
outputs of the teacher as hard targets.

\noindent{\bf Stopping rules.}
In addition to establishing a timeout for distillation, \approach
can terminate  distillation when the relative
student-teacher error is below a given threshold.
For example, in a regression setting, the MSE between the
teacher and student outputs may be used to measure relative accuracy.

Once parameterized, \approach trains the student network automatically.
The resulting network can be assessed for accuracy and subjected to
property verification.

\subsection{Implementation}

Our implementation of \approach consists of two main components:
a transformation component, and a distillation component. Both components
are parameterized using a single configuration file, specified using the TOML
language~\cite{toml}, such as the one
shown in Listing~\ref{lst:config}.

\begin{figure}
\begin{lstlisting}[
        language=toml,
        frame=single,
        numbers=left,
        numberstyle=\tiny\color{mygray},
        xleftmargin=3em,
        framexleftmargin=2em,
        caption=An example \approach configuration file.\label{lst:config}
    ]
[distillation]
maxmemory="32G"
threshold=0.01
type="regression"
[distillation.parameters]
epochs=100
optimizer="adam"
loss="MSE"
[distillation.data]
format="dronet"
batchsize=32
[distillation.data.transform]
grayscale=true
[distillation.data.train]
shuffle=true
student.path="artifacts/dronet.200/training"
teacher.path="artifacts/dronet.200/training"
[distillation.data.validation]
shuffle=false
student.path="artifacts/dronet.200/validation"
teacher.path="artifacts/dronet.200/validation"
[distillation.teacher]
model="networks/dronet/model.onnx"
input_shape=[1, 200, 200, 1]
input_format="NHWC"
[[distillation.strategies.drop_layer]]
layer_id=[2, 3]
[[distillation.strategies.forall]]
layer_type="ResidualConnection"
strategy="linearize"
[[distillation.strategies.scale_layer]]
layer_id=[0]
factor=0.5
\end{lstlisting}
\end{figure}

The transformation component is parameterized with a teacher architecture and
a set of transformation strategies. The teacher model is specified using the
Open Neural Network Exchange~(ONNX) format~\cite{onnx},
which can be produced from many of the most popular machine learning frameworks.
The configuration file in Listing~\ref{lst:config} will transform the teacher
architecture (specified in lines 22-25) by dropping layers 2 and 3
(lines 26-27), linearizing all residual blocks in the network (lines 28-30),
and scaling the first layer by a factor 0.5 (lines 31-33). After
transforming the architecture, the \approach tool will automatically begin
distillation.

The distillation component was implemented in Python, with PyTorch as the
underlying machine learning framework~\cite{pytorch}
used for training. The default distillation settings are based on the knowledge
distillation method of Hinton~et~al.~\cite{hinton-etal:distillation}, and can
be configured through the configuration file provided by the user. 
\approach allows users to choose the training algorithm and loss function (lines
7 and 8), as well as set hyper parameters such as the learning rate, 
and the batch size (line 11). We also extend distillation with a 
relative-performance-based stopping criteria (such as the error threshold 
specified in line 3), as well as a memory limit (line 2) and timeout. 
These permit developers to enforce a distillation budget 
to allow for fast partial distillation training to rapidly explore the space of refactorings; something we plan
to explore in more depth in future work.
Our distillation procedure also allows for different datasets to be specified
for the student and the teacher (lines 14-21). This allows the student to 
be trained on smaller input sizes or on inputs that have been pre-processed 
differently than the teacher. For example, in the second case study below,
we transform the \dronet network to operate on smaller input sizes, so we
update lines 16 and 20 to point to a pre-processed dataset with the smaller
image sizes.
The distilled student model is saved in the ONNX format.

We also developed several tools for translating
the ONNX models to the input formats required by the verifiers used in this work. 
\S\ref{sec:study} provides  details on these translators. 

We will make our implementation and study artifacts available before the work is published.

\subsection{Recommended Practices}
\label{sec:practices}
Our experience to date with \approach has led to several broad,
and likely evolving, recommended practices to explore the 
space of DNN refactorings possible with the approach.

First, the configuration language provides a simple and unified way to specify architectural transformations, and \approach removes potential errors associated with the transformation itself by automating this process. Still, when specifying architectural transformations, a developer may be challenged to identify the complexity sweet spot given the possible number of applicable transformations. 
We recommend to initially use a binary search to explore the complexity spectrum, starting with the original network and applying coarse \textit{drop} transformations to generate networks with half the complexity and determine whether the accuracy and verifiability rendered are acceptable. 
Once the sweet spot is identified, further transformations can be applied to refine this initial approximation.  For example, \textit{scale} transformations 
can provide finer gradations in DNN complexity than using \textit{drop} alone.
We explore this recommendation further in \S\ref{sec:results}, and certainly a key part of future work is to provide further support for this search process including making the binary search process multidimensional to account not just for the number of layers to drop but also for the layer type and its location in the DNN. 

Second, the configuration language  provides a means to specify how the distillation process should work, including training and validation data sets, loss functions, and training parameters.  
With regard to datasets for \approach, we advocate a strategy
of exploiting, whenever available, datasets used for the original  network. The intuition here is that the knowledge learned
by the teacher, has been informed by
repeated application of training data from those sets and distilling will effect the transfer of that well-informed knowledge to the student.

Third, the sweet spot is likely to vary across verifiers. As described in \S\ref{sec:overview}, many verifiers are emerging with different underlying approaches that may present performance and applicability tradeoffs, and it is unlikely that a single best verifier across the space of DNNs and properties will emerge.  To address such
richness,  we recommend an \approach portfolio approach, 
where an ensemble of strategies can be employed
to generate a portfolio of refactored networks, each of which can then
be subjected to verification by a portfolio of verifiers.
Selection of the best refactored DNN, e.g., the most accurate verifiable 
network, can be made from the portfolio results.
Furthermore, such refactoring and verification portfolios would be
comprised of independent
tasks which can be easily parallelized -- a practice 
employed in conducting the case studies discussed in \S\ref{sec:study}.

\section{Study}\label{sec:study}

This section sets up the three case studies we conduct to explore the potential
of \approach to support DNN verification.

The first case study focuses on Verifier Applicability.
Verifiers are constantly playing catch-up to the latest architectural features introduced for DNNs.
As a result, it is common to find that a favorite verifier does not 
support the features of a network of interest.
For example, some verifiers, such as CROWN and ReluVal, are restricted to fully connected layers.
Other verifiers, such as MIPVerify and DeepGo, add support for convolutional and maxpooling, but 
cannot handle more complex layers such as residual blocks.
In this case study, we showcase how \approach can be used to overcome this kind of limitation by refactoring
the network to one that satisfies the verifiers feature constraints.

The second case study focuses on Verifier Efficiency. 
When faced with modern complex networks, even when verifiers can support all the features in a network,
they often struggle to provide results in actionable time.  In such cases \approach
can be used to refactor the network into one that can be verified significantly more efficiently.

The third case study focuses on Error-Verifiability Trade-offs. 
While the first two studies investigate verification applicability and efficiency enabled by \approach
under a small number of refactorings, this study provides a much broader exploration of the refactoring
space. 
It does so by performing a binary search for refactored versions of the original DNNs that
lie in the \textit{complexity sweet spot} where error and verification
time meet acceptable thresholds, and then it follows up with a more extensive 
characterization providing concrete evidence of the tradeoffs illustrated earlier in the paper.

The studies are performed on two large DNN artifacts,
four state-of-the-art verifiers, and multiple refactored DNNs
in order to provide a broader characterization of the approach.
Table~\ref{table:studies} summarizes the case studies and how they pair networks and verifiers.

\begin{table}
	\centering
	\caption{Case Studies}
	\begin{tabular}{lll}
		\textbf{Case Study}            & \textbf{Artifacts} & \textbf{Verifier} \\
		\toprule
		Verifier Applicability         & \dave              & \reluplex         \\
		                               & \dronet            & \eran             \\
		\midrule
		Verification Efficiency        & \dave              & \neurify          \\
		                               & \dronet            & \planet           \\
		\midrule
		Error-Verifiability Trade-offs & \dave              & \neurify          \\
		\bottomrule
	\end{tabular}
	\label{table:studies}
\end{table}

\subsection{Artifacts}

The study applies \approach to two large DNN artifacts used for autonomous vehicles, \dave and \dronet,
both with more than 10 layers of different types, and  an average of 250,000 neurons respectively.
Prior DNN verification efforts have typically focused on relatively simple DNNs (e.g., ACAS has 8 layers and hundreds of
neurons~\cite{katz-etal:CAV:2017}, MNIST has four layers and thousands of neurons~\cite{ehlers:ATVA:2017})
or manually simplified versions of larger DNNs (e.g., a version of DAVE-2 with half of the 
layers~\cite{wang-etal:NIPS:2018:neurify}) to assess their performance.
Our goal is to explore the performance of \approach
under modern and challenging DNNs that push the state-of-the-art.

\vspace{0.1in} \noindent \textbf{DAVE-2} is trained to drive an autonomous ground vehicle.
The \dave architecture~\cite{bojarski-etal:corr:2016:DAVE2} is made up of over 80000 neurons
spread over 13 layers, including five convolutional layers and four fully connected layers, and
was designed to provide steering control for an autonomous car.
Because the original trained \dave model is not publicly available, in this
study we use the model trained by other researchers~\cite{pei-etal:SOSP:2017:deepxplore}
with the Udacity self-driving car
dataset~\cite{udacity-sdc}. The
dataset is comprised of 101,396 training images (of which we use 90\% for
training, and 10\% for validation) and 15837 test images. Images are taken
from three cameras (center, left, right) looking out the windshield of a
car. For each image, there is also a corresponding steering angle, indicating
the position of the steering wheel at the time the image was taken.
The \dave model that we use was trained by Pei et al.~\cite{pei-etal:SOSP:2017:deepxplore} on color images that were scaled
down to a height and width of 100 by 100. We use the same input pre-processing when refactoring
the network. The resulting network has a MSE of 0.008.

Following the process of \neurify~\cite{wang-etal:NIPS:2018:neurify},
we randomly generated 10 safety properties for the \dave networks. The
properties specify that images with a Chebyshev distance of at most 2 from a given test image must
produce an output steering angle within a target range. Different from \neurify~\cite{wang-etal:NIPS:2018:neurify},
in order to make the properties independent of an individual network,
we use the test data label as the target steering angle. Another difference  
is that we restricted the output to a tighter range (from $\pm30$ to $\pm10$ degrees)
to make it more realistic for the problem domain.

\vspace{0.1in} \noindent \textbf{\dronet} is
trained to provide control signals for an autonomous
drone~\cite{loquercio-etal:RAL:2018:dronet}.
\dronet is based on the ResNet architecture, and is made up of
a convolutional and maxpooling layer, followed by 3 residual blocks, and
a fully connected layer, totalling over 475,000 neurons. It consumes a 200x200 grayscale
image and outputs a steering angle, as well as a probability that the drone is
about to collide with an obstacle. The model is trained on a modified version
of the Udacity self-driving
car dataset. The dataset only uses the images from the center camera, and their
corresponding steering angles from the Udacity dataset. It augments these
examples with a set of images taken from a camera mounted to a bicycle as it
is ridden on the streets of a city. These images each have a label indicating a
corresponding probability of collision. The resulting network has a MSE of 0.021 
on steering angle and an accuracy of 0.983 on collision avoidance.

We randomly generated 20 safety properties for the \dronet networks, 10 for
the steering angle regression task, and 10 for the collision probability classification
task. Similarly to the properties for \dave, the \dronet steering properties specify
that images within a Chebyshev distance of 2 of a given test image must produce a steering angle within
10 degrees of the steering angle of the original. The collision probability properties
specify that images within a Chebyshev distance of 2 of a given test image must produce a collision
classification that is the same as the target classification.

\subsection{Verifiers}

Given our studies' objectives, we selected the following verifiers \reluplex, \eran, \neurify, and \planet.
\reluplex and \planet are the most studied DNN verifiers, while
\eran and \neurify represent the state-of-the-art.  
All these verifiers also have freely available open source implementations.
The \reluplex verifier is based on an adaptation of the simplex method that provides
support for DNN architectures with fully-connected layers with ReLU activation functions.
The \eran verifier uses abstract interpretation to overapproximate the reachable
output region. In this study we use \eran with the DeepPoly abstract domain, which
can handle convolutional, fully-connected, and maxpooling layers with ReLU, tanh, or
sigmoid activation functions~\cite{singh-etal:POPL:2019:deeppoly}.
The \neurify verifier combines symbolic interval analysis with symbolic linear relaxation,
and supports convolutional and fully-connected layers with ReLU activations.
The \planet verifier blends techniques from SMT and search, and supports layers that can
be specified as a linear combination of neurons, as well as ReLU activations and maxpooling layers.

For each verifier, we wrote a translation tool to convert networks in the ONNX format
(the output format of \approach) to the specific input format required by each verifier.

Both \reluplex and \planet take in a single file that combines a description of the
network and the property specification. In the `nnet' file format of \reluplex, the property
is encoded as a set of layers on the end of the network. The types of properties supported by this format
are ones in which the input constraints are intervals, and the output constraint is a linear
inequality over the output neurons.
In the `rlv' format used by \planet,
layers of a network are specified as linear combinations of neurons, and the property is 
specified as a list of linear assertions, with each input required to have a lower and upper bound. 
We wrote translation tools to combine an ONNX network and property into the corresponding file type.
For \reluplex, we support translating networks comprised only of fully connected layers.
For \planet, we support translating any networks comprised of fully connected and convolutional layers
with ReLU activations, as well as maxpooling, batch normalization, and residual blocks.

Rather than combining the network and property into a single file, \neurify and \eran take in both 
a network specification and several parameters to specify the property to be checked, such as the 
input and output bounds.
The extended `nnet' format used by \neurify accepts networks composed of convolutional and fully
connected layers that must have ReLU activations.
The \eran verifier accepts networks specified as a subset of the operations in the TensorFlow
machine learning framework.
We built translation tools to convert the ONNX models output by \approach to the network
specification format accepted by each verifier.
For \neurify, we support translating networks comprised of convolutional and fully connected layers.
For \eran, we support fully connected and convolutional layers
with ReLU or sigmoid activations, as well as batch normalization layers, by converting them to
fully connected layers.

For some of the verifiers we had to make small modifications to get the tool to work on the
artifacts used in this study. For \reluplex, we use the version of the tool modified 
by~\cite{DBLP:conf/nips/BunelTTKM18} to support
generic properties. This version of the tool works by checking whether the output neuron is
less than 0. Properties can be specified by encoding them as a set of layers at the end of
the network being verified.
Because the \neurify tool is hard coded to check a constant set of properties, we modified 
the version of the tool built to operate on \dave for use in our study. We made the tool more general by allowing
a user specified input and output interval to be specified at runtime.
Finally, we modified \eran to support regression networks. The original tool was designed
to verify robustness properties for classification networks. We made modifications to the
tool to return the lower and upper bounds that it had computed for the outputs. We then
checked whether these bounds were within the bounds specified by the property.

\subsection{Methodology}

For  each case study, we use \approach to refactor
the original DNN. The transformations in
each scenario are meant to  demonstrate how \approach can be applied
to increase verifiability.
Distillation parameters are set to correspond to the
training parameters reported for  the original \dave and \dronet,
including
the number of epochs we used (50 for \dave and 100 for \dronet), 
the optimization method (Adadelta  for \dave and Adam for \dronet),
and the batch size (256 for \dave and 32 for \dronet).
After training, we select the model
from the epoch with the best mean squared error (MSE) performance (relative to the
teacher), evaluated on the validation set. 
For \dave, the performance is measured as the MSE with respect to the target steering
angle.
Because \dronet has 2 outputs, we report the performance as 2 values: the MSE with respect
to the target steering angle, and the accuracy with respect to the probability of collision.

For each scenario in each case study, we use the corresponding verifier to check the
safety properties described above on the corresponding refactored network. We measure
the verification time and report the verification results (true, false, unknown, ``out of resources'' (OOR))
as well as the mean and standard deviation of the verification time.

Distillation took around 11 hours on average to complete all the epochs. Note that we made no effort
to optimize this time, such as using early stopping or transfer learning.
As an example of the potential for optimization,
the average best epoch for the accuracy-verifiability trade-off study below was 37 (out of 50 total).  
This suggests that simply using
early stopping would have reduced distillation time by 20\%.
More aggressive training optimizations, such as those used in
neural architecture search~\cite{pham-etal:icml:18}, apply a strict
training budget, e.g., 5\% of the epochs, when making preliminary
assessments about the suitability of an architecture and then only
using larger budgets for very promising architectures. 
This would work well for the approach described in Case Study III and
would reduce the time for most distillation runs 
to a matter of minutes.

Distillation was run on GPU computing nodes with 64GB of memory, and using
NVIDIA 1080Ti GPUs for network training.
Verification tasks were run on compute nodes with 2.3GHz Xeon processors
and 64GB of memory running CentOS Linux 7.
Verification tasks had a time out of 24 hours.

\section{Results}
\label{sec:results}

In this section, we present the results of each case study.

\begin{table}[t]
	\centering
	\caption{Results for Case Study I. }\label{tab:cs1}
	\begin{tabular}{l|rrrr|rrr}
		                           & \multicolumn{4}{c|}{\textbf{Property}}               & \multicolumn{3}{c}{\textbf{Mean Verif.}}                                           \\
		                           & \multicolumn{4}{c|}{\textbf{Count}}                  & \multicolumn{3}{c}{\textbf{Time (sec.)}}                                           \\
		\textbf{Artifact/Verifier} & T                                                    & F                                        & U & \rotatebox{90}{OOR} & T   & F   & U \\
		\toprule
		\dave/\reluplex            & \multicolumn{7}{c}{\dave Not Supported by \reluplex}                                                                                      \\
		R4V(\dave)/\reluplex       & 10                                                   & 0                                        & 0 & 0                   & 52  & -   & - \\
		\midrule
		\dronet/\eran              & \multicolumn{7}{c}{\dronet Not Supported by \eran}                                                                                        \\
		R4V(\dronet)/\eran         & 8                                                    & 12                                       & 0 & 0                   & 519 & 525 & - \\
		\bottomrule
	\end{tabular}
\end{table}

\subsection{Verification Applicability}

\vspace{0.05in} \noindent {Artifacts and Verifiers.}
This case study pairs the \dave and \dronet artifacts with the \reluplex and \eran (using the DeepPoly abstract domain)
verifiers. The pairing is designed to showcase the potential value of \approach to overcome a verifier's limited support for certain DNN features.
The original \dave network has several convolutional layers, which are not supported by \reluplex.
The original \dronet network has complex residual blocks. \eran, however, does not support residual blocks
when using the DeepPoly domain (indeed this increasingly common layer type is only supported by
a limited set of verifers, e.g., CNN-Cert).

\vspace{0.05in} \noindent{Refactoring.}
Since \reluplex only supports fully-connected layers, \dave must be refactored to remove all its convolutional layers.
The refactoring uses the  transformation $drop(\{0,1,2,3,4\})$ to remove all unsupported layers, leaving only the final 5 fully-connected layers.
To enable \eran (DeepPoly) to run on \dronet, we first drop the convolutional and
maxpooling layers at the beginning of \dronet, as well as the last 2 residual
blocks through $drop(\{0, 2, 3\})$.
We then  linearize
the remaining residual connection $linearize(\{1\})$.
The distillation for both artifacts uses the parameters described in the previous section,
resulting in a refactored \dave  with an MSE of 0.062 relative to the original (other refactorings of \dave
that render much smaller errors are presented in the last case study)
, and a refactored \dronet
with a relative MSE on the steering angle of 0.020 and a relative accuracy on the collision avoidance of 0.979.

\vspace{0.05in} \noindent{Findings.}
The results from this case study are shown in Table~\ref{tab:cs1}. The first column
of this table is the paired artifact and verifier, the next four columns are the number of properties
that were checked that resulted in either proving or falsifying the property,  returning
unknown, or running out of resources. The next three columns show the mean   verification
time across all properties for each result type.

While \reluplex can not be applied to the original \dave artifact, and \eran (DeepPoly) could not
be applied to the original \dronet artifact, both verifiers could be applied to the
refactored networks.
In the case of \dave and \reluplex, all 10 properties where shown to hold in less than a minute per property.
In the case of \dronet and \eran (DeepPoly), the refactoring enabled the verification of all 20 properties,
eight deemed true and 12 deemed false, taking on the order of 10 minutes to complete the verification of each property.

\textbf{The most significant finding is that \approach  enabled the application of previously inapplicable verifiers.}

\begin{table}[t]
	\caption{Results for Case Study II}\label{tab:cs2}
	\begin{tabular}{l|rrrr|rrr}
		                           & \multicolumn{4}{c|}{\textbf{Property}} & \multicolumn{3}{c}{\textbf{Mean Verif.}}                                                    \\
		                           & \multicolumn{4}{c|}{\textbf{Count}}    & \multicolumn{3}{c}{\textbf{Time (sec.)}}                                                    \\
		\textbf{Artifact/Verifier} & T                                      & F                                        & U  & \rotatebox{90}{OOR} & T    & F     & U      \\
		\toprule
		\dave/\neurify             & 0                                      & 0                                        & 10 & 0                   & -    & -     & 20318* \\
		R4V(\dave)/\neurify        & 10                                     & 0                                        & 0  & 0                   & 2449 & -     & -      \\
		\midrule
		\dronet/\planet            & 0                                      & 0                                        & 0  & 20                  & -    & -     & -      \\
		R4V(\dronet)/\planet       & 1                                      & 2                                        & 0  & 17                  & 1155 & 45423 & -      \\
		\bottomrule
	\end{tabular}
\end{table}

\subsection{Verification Efficiency}

\begin{table*}[ht]
    \centering
    \caption{Case Study III: Searching for Viable Architectures with \approach. }\label{tab:cs3}
    \begin{tabular}{llrrcc}
        \textbf{Search Action} &  \textbf{Network}    & \textbf{\#Neurons} & \textbf{Rel.MSE} & \multicolumn{1}{c}{\textbf{Verification}} & \multicolumn{1}{c}{\textbf{Assessment}}\\
        \toprule
        Start with existing network & Original            & 82669               & 0.0   & 10/10 Unknowns & Verification returns too many \\ 
        & & & & Over 3 hours per property &  unknowns and takes too long  \xmark \\
                \midrule
        Drop all but one computation   & \_\_\_\_\_56\_\_\_A & 11  & 0.062  & Verified 10/10 in seconds & Quickly verifiable but   \\
        layer & & & & & error is too large  \xmark\\
        \hline 
        Drop 4 layers (convolutional) &      0\_\_\_\_56789A     & 56621     & 0.008    & Verified 10/10 in less than 3 hours  & Verifiable but error is borderline    \cmark        \\
        Drop 4 layers (fully-connected) &        0123456\_\_\_\_      & 81345     & 0.008   & Verifier cannot process  & Incompatible with verifier    \xmark   \\
        \hline
        Drop 2 layers (convolutional) &        012\_\_56789A       & 77933       & 0.005    & Verified 6/10, 3 Unknowns, 1 TO  & Partially verifiable  \\ 
                & & & & Under 3 hours per property &   and acceptable error   \xmark  \\
        Drop 2 layers (fully-connected) &      0123456\_\_9A       & 81405     & 0.003  & Verified 10/10  & Verifiable and             \\
                       & & & & Just under 3 hours per property &  acceptable  error        \cmark  \\
        \bottomrule
    \end{tabular}
\end{table*}

\vspace{0.05in} \noindent {Artifacts and Verifiers.}
This case study pairs \dave and \dronet with the \neurify and \planet verifiers.
These verifiers can run these networks, but on the original architectures
\neurify takes  on average over 5.6 hours to check a single property
while \planet cannot verify any \dronet property within 24 hours.

\vspace{0.05in} \noindent{Refactoring.}
To improve the efficiency of \neurify on the \dave network, we  apply \approach
to refactor \dave by applying the   transformation $drop(\{2, 3, 4, 7, 9\})$,
which matches the one manually generated by ~\cite{wang-etal:NIPS:2018:neurify}.
To increase the efficiency of \planet on \dronet, we apply \approach
to remove the first convolutional and maxpooling layers, as well as the last two
residual blocks ($drop(\{0, 2, 3\})$). We also scale the size of the input layer to a 10 by 10
grayscale image ($scale(\{input\}, 0.05)$). This last step is necessary to enable planet to verify any
of the properties within the time limit.

\vspace{0.05in} \noindent{Findings.}
The results from this case study are summarized in Table~\ref{tab:cs2}. The first column
of this table is the artifact and verifier pair, the next four columns are the number
of properties that were checked that resulted in either proving or falsifying the
property, or returning unknown or running out of resources. The next three columns
show the mean  verification time for each kind of returned result.

While \neurify could mostly handle the original \dave network, its results were flaky
for some properties, alternating segmentation errors with verification completions.
Hence, the reported numbers on the original \dave with \neurify correspond to the best of
up to five tries.
But even when \neurify completed its execution after an average of 339 minutes per
property,  the complexity of the original network introduced significant
imprecision into the analysis, causing all 10 property checks to return
an \textit{unknown} result. In contrast, \neurify was able to return true
for all properties of the refactored network, taking about 41 minutes on average
to verify a property - an 8 times speedup.

Similarly, although \planet can accept the original \dronet architecture, it
could not complete any of the property checks within one day. After applying
\approach to refactor the network, \planet was able to finish checking three out
of the 10 properties, taking just under 20 minutes to verify a property and
a little over 16.5 hours to falsify a property.

\textbf{The most significant finding is that \approach can speed up the verification time
	and enable verifiers to provide more useful results.}

\subsection{Accuracy and Verifiability Trade-offs}

\vspace{0.05in} \noindent {Artifacts and Verifiers.}
In this case study, we explore the error-verifiability trade-offs using the
\dave architecture and the \neurify verifier.
We generate multiple refactored versions of \dave to explore the trade-offs.

\vspace{0.05in} \noindent {Refactoring.}
We start our exploration of the refactoring space using the coarser granularity transformation \textit{drop}, and
performing a binary search to determine the number of layers to drop based on the
refactored network error and verification time. We arbitrary set the acceptable
error in terms of the $MSE$ relative to the original network to be under 0.01.
In terms of verifiability, we require for all the properties to be verified as true
or falsified in under three hours.

\begin{figure*}[ht]
	\centering
	\includegraphics[width=\linewidth]{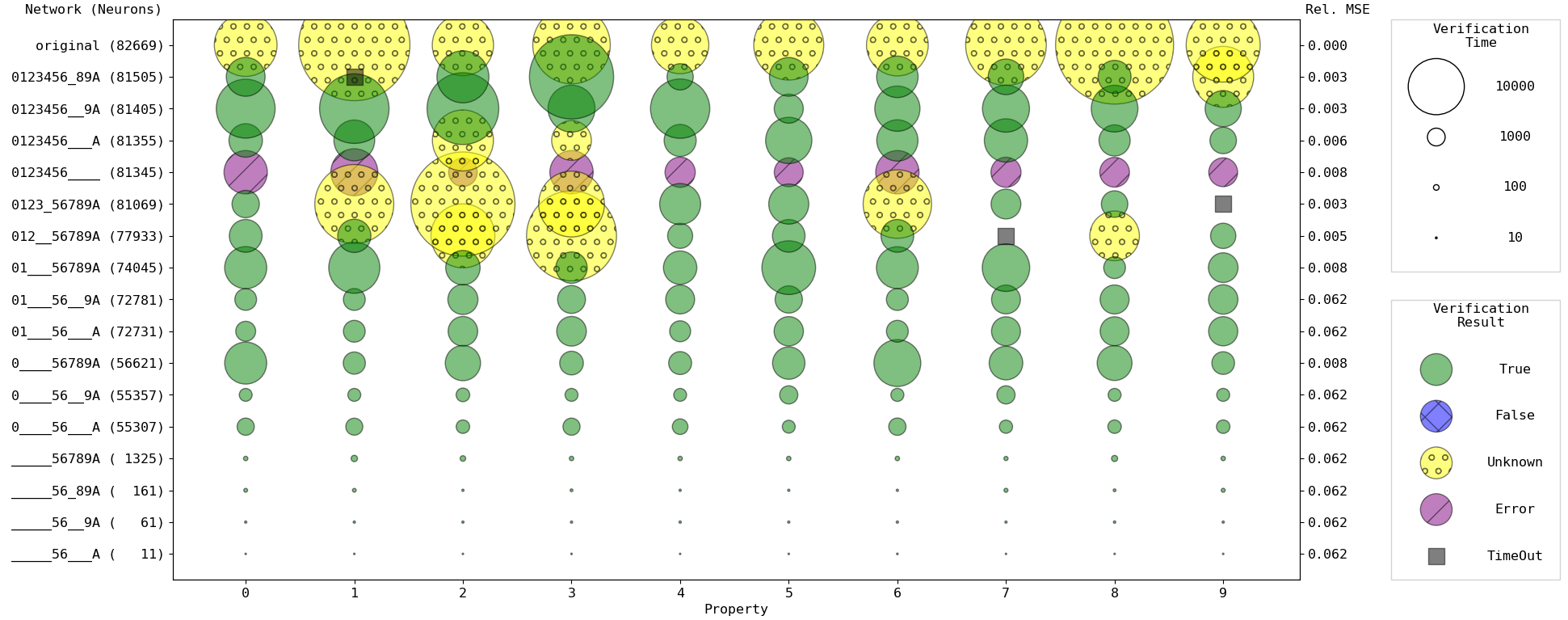}
	\caption{Results for Case Study III}\label{fig:cs3}
\end{figure*}
\vspace{0.05in} \noindent{Findings.}
The binary search process we followed is summarized in Table~\ref{tab:cs3}.
The first column describes the search action taken in terms of the refactoring operation.
The second column shows the name of the network.
In \dave the first five layers are convolutional layers,
layers five and six are reshaping layers,
which cannot be modified by \approach, and the last four layers
are fully-connected.
The name of each network includes all of the layers that are kept after transformation, and
replaces the names of dropped layers with an underscore.
The third column shows one measure of the network complexity
in terms of the number of neurons. The fourth column shows the relative mean square error
when compared with the original network. The fifth column provides a summary of the verification
outcome (more details are provided next). The last column contains the assessment to guide
the search process.

The process shows how a systematic binary search, starting from the original
network and repeatedly bisecting the space towards the direction that seems to offer the best
tradeoffs, renders a refactored network 0123456\_\_9A that performs according the target criteria.
This network can verify all 10 properties
in less than three hours per property and do so while retaining
the behavior of the original network with a MSE under 0.01.
The process also hints at how even when considering a single transformation, many options
arise for exploration and refinement, including some options that we have illustrated
and others that we have yet to explore such as the  combination of \textit{drop} with \textit{scale} transformations.

To complement the data in  Table~\ref{tab:cs3} and  to provide a more extensive characterization
of the space of tradeoffs involved we use  \approach to refactor the original network into
another 11 networks. The results are depicted in Fig.~\ref{fig:cs3}.
Each row in this figure is a refactored network, and each column is a property to verify.
The networks names and their number of neurons appears on the first and second column.
The networks are sorted by the number of neurons, so networks towards
the bottom of the plot are simpler.
The size of the circle represents the time to verify a given network and property, and
the color and pattern of the circle indicate the verification result.
The relative MSE of each network appears on the right of the plot.

Starting from the top, we notice that the original \dave network takes over 10,000 seconds to verify
every single property, and for the second property as much as 39306 seconds,
and in all cases it returns \textit{unknown}.
When \approach is applied to refactor \dave by removing the fully-connected layer labeled as  '7'
resulting in 0123456\_89A
(second row in the graph),  we observe  a 2.1x speedup in
verification time and also more
useful returned results (eight properties were shown to be true, one unknown, and one timed-out).
The third row shows network 0123456\_\_9A, which has  another fully-connected layer removed, resulting in a
speedup of 3.6x in verification time and all properties checked as true.
The following two rows of networks keep dropping fully-connected layers, and when all fully-connected
layers are dropped we notice that \neurify returns an error as it cannot deal with resulting architecture
of 0123456\_\_\_\_ which is correct but has no fully connected layers, which was unexpected by \neurify.
The following rows start the removal of convolutional layers, resulting in another noticeable reduction in
network complexity. Network 01\_\_\_56789A (highlighted in \S\ref{sec:introduction}) renders a speedup of 6.3x
in verification time while retaining a  MSE of 0.008 relative to the original network.
That is also the case with 0\_\_\_\_56789A, with 9.4x speedup and still an acceptable error rate.
The rows below have too large of an error to be deemed viable.

In total, 3 of the 16 refactored \dave networks in Fig~\ref{fig:cs3} meet
the error-verifiability criteria, suggesting that the complexity sweet
spot for \dave contains a variety of acceptable DNNs.

\textbf{The most significant finding of this study is that, without prior domain knowledge,
	\approach can enable the effective exploration of the error and verifiability tradeoffs and
	converge towards a network that meets both criteria}.

\section{Conclusion}\label{sec:conclusion}

In this work, we proposed automated support for refactoring DNNs to 
facilitate verification while retaining the behavior of the original network.
\approach couples specification-driven architecture transformation support with
knowledge distillation to produce refactored DNNs.
We demonstrate, through a series of case studies, that \approach can help
developers overcome the limitations of individual DNN verification tools, reduce
the time to produce verification results, and, most importantly, to 
navigate the accuracy-verifiability tradeoff to arrive at a network
that meets stated accuracy requirements and verification budgets.
These findings suggest that \approach has the potential for improving the
development of critical systems involving DNNs, since refactored DNNs
can be subjected to property verification and have acceptable accuracy
for deployment.

In future work, we plan to expand the range of refactorings, e.g.,
adding \textit{add} and \textit{replace} operators, 
extend existing operations, e.g., linearization of dense networks,
and supporting additional optimizations, e.g., transferring weights 
from the original network, using inexpensive training parameters
during the search for a good refactoring.
In addition, we plan to explore different strategies for
automating the search
for refactored DNNs that lie within the complexity sweet spot, which
has the potential to make \approach broadly applicable by practitioners.

\section*{Acknowledgment}
This material is based in part upon work supported by the 
National Science Foundation under Grant Number 1617916, 1900676, and 1901769,
and the U. S. Army Research Laboratory 
and the U. S. Army Research Office under contract/grant number W911NF-19-1-0054-P00001.

\bibliographystyle{IEEEtran}
\bibliography{references}

\end{document}